\documentclass[10pt]{article} % For LaTeX2e
\usepackage[preprint]{tmlr}
\usepackage{amsmath,amsfonts,bm}

\def\eqref#1{equation~\ref{#1}}
\def\1{\bm{1}}

\DeclareMathAlphabet{\mathsfit}{\encodingdefault}{\sfdefault}{m}{sl}
\SetMathAlphabet{\mathsfit}{bold}{\encodingdefault}{\sfdefault}{bx}{n}

\usepackage{hyperref}
\hypersetup{hidelinks}
\usepackage{url}
\usepackage{graphicx}
\usepackage{booktabs}
\usepackage{placeins}
\usepackage{float}

\title{Decodable but Not Actionable: \\Auditing the Representation–Behavior Gap in Frozen LLMs for Geometric Constraints}

\author{\name Man Liang \email liang558@umd.edu \\
      \addr University of Maryland
      \AND
      \name Xinzhao Cheng \email xzcheng@umd.edu \\
      \addr University of Maryland
      \AND
      \name Faizan Wajid \email fwajid@umd.edu\\
      \addr University of Maryland}

\def\month{MM}  % Insert correct month for camera-ready version
\def\year{YYYY} % Insert correct year for camera-ready version
\def\openreview{\url{https://openreview.net/forum?id=XXXX}} % Insert correct link to OpenReview for camera-ready version

\begin{document}

\maketitle

\begin{abstract}
Large language models (LLMs) perform well on structured reasoning
tasks, but what they encode and whether encoded information guides
their behavior remain unclear. We introduce a controlled auditing
framework that uses parametric CAD to distinguish failures of
encoding, expression, and activation-level control. Applied to eight
pretrained models and matched randomly initialized controls, our
framework reveals that pretraining contributes substantially more to
linear decoding of local relations than to global constraint status,
which is already highly decodable from random representations. Yet
strong local decodability does not translate into the models'
zero-shot predictions, which remain far below supervised probe
performance. Restoring clean internal activations can repair probe
predictions, whereas simple activation steering does not provide
reliable targeted control. With only 400 labeled relation pairs,
low-rank fine-tuning substantially improves generation on three
representative models, showing that the representation--behavior gap
can be reduced through adaptation. Together, these findings show why
decodability alone is insufficient evidence that encoded information
will be expressed or controllable in model behavior.\footnote{AI tools were used to assist with manuscript preparation.
The research ideas, experimental design, analyses, and conclusions are
those of the authors. All AI-assisted material was reviewed and
verified by the authors.}
\end{abstract}

\section{Introduction}

Large language models are increasingly used to translate natural
language instructions into structured outputs such as programs, plans,
and geometric designs
\citep{wuAutoGenEnablingNextGen2023,
singhProgPromptGeneratingSituated2023,
wuChat2SVGVectorGraphics2025}. For such systems to be trustworthy in
scientific and industrial applications, their outputs must satisfy
interacting domain constraints while remaining globally consistent.
Parametric CAD provides a useful setting for studying this problem
because it is widely used in engineering and represents constraints in
an explicit, verifiable form~\citep{seff2020sketchgraphs}. Local
constraints specify relations between geometric elements, while their
combined effect determines whether a sketch is underconstrained, well
constrained, or overconstrained
\citep{thierryExtensionsWitnessMethod2011}.

Recent work has progressed from generating CAD designs directly from
natural language instructions~\citep{khan2024text2cad,
liCADLlamaLeveragingLarge2025} to iterative systems that use execution
or solver feedback to detect and repair errors
\citep{fanTraceCADTraceGuidedRepair2026,
huIterCADIterativeMultimodal2026,
liuEmbodiedCADSolverGrounded2026}. However, evaluating outputs alone
does not reveal whether successful generation and repair rely on
internal representations of geometric constraints. This distinction
may explain why competence on local relations can coexist with failure
at the sketch level. We therefore ask what geometric constraints frozen LLMs encode,
whether this information is expressed in generation, and whether
activation interventions can influence or control model behavior.

To answer this question, we develop a four-part audit of linear
decodability, forced-choice generation, activation-level influence,
and steering. We apply the framework to eight pretrained language models spanning Qwen2.5, Qwen3.5, Mistral, Llama-3.1, and
OLMo3~\citep{qwenQwen25TechnicalReport2025,jiang2023mistral,
grattafiori2024llama3herdmodels, qwen3.5, olmoOlmo32026} on SketchGraphs~\citep{seff2020sketchgraphs}. The framework includes three matched evaluation tasks: local decoding of pairwise constraints (P1), global decoding of degrees-of-freedom status (P2), and forced-choice generation using the same pairwise labels as P1 (P3). All tasks use
geometry-only serializations from which constraint annotations are
removed. To characterize the P1--P3 gap beyond correlational probing, we apply
activation patching and mean-difference steering to three
representative models, testing whether decoded relation information
affects probe predictions and permits targeted control. We then use
low-data LoRA to determine whether the generation deficit can be
overcome with limited supervision.

Our results reveal a systematic gap between decodable and behaviorally
available geometric constraint information. Local pairwise relations
(P1) show a consistent pretraining-specific gain over matched
randomly initialized representations, and part of this gain remains
after entity-order shuffling. Global constraint status (P2), however,
benefits only modestly from pretraining. This local-to-global
asymmetry is captured by a dissociation index that is positive across
all eight models. More importantly, strong P1 decodability does not
translate into reliable P3 generation, with failure severity varying
sharply across model backbones. Activation patching on three
representative models shows that restoring clean entity activations
repairs probe predictions most strongly in early layers. The effect
declines with depth but remains measurable at the probe readout layer.
Mean-difference steering does not reliably control predictions, while
limited LoRA supervision substantially improves relation prediction. This distinction matters because poor task performance alone cannot
determine whether relevant information is weakly represented or not
expressed in the tested behavior. Our controlled, cross-model framework
addresses this ambiguity by separating evidence for representation from
evidence for expression and control, and by testing whether supervision
can reduce the gap between them.

\begin{figure}[t]
\centering
\includegraphics[width=\linewidth]{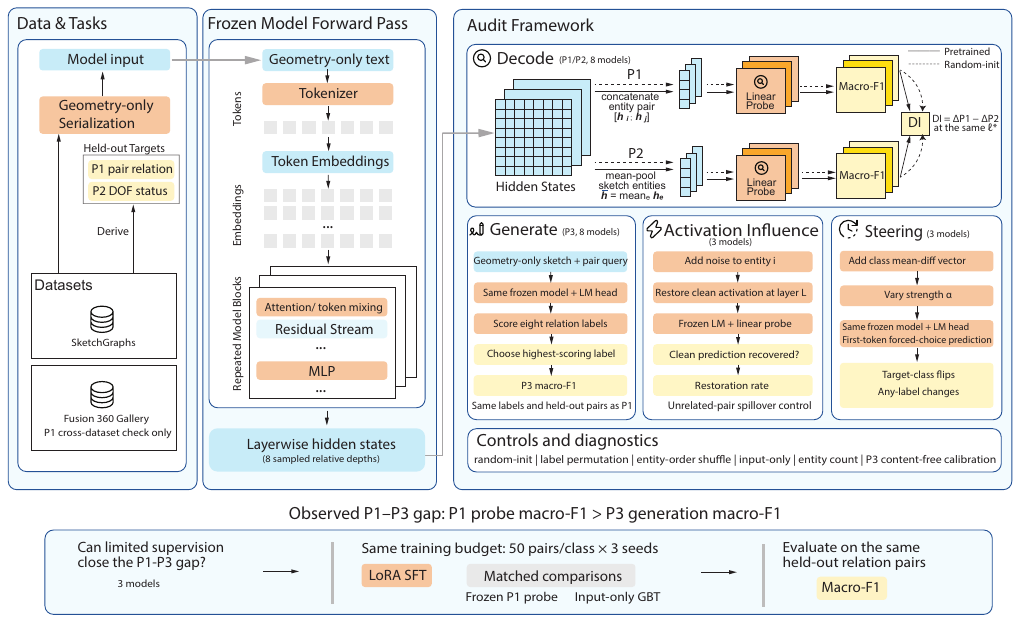}
\caption{Overview of the audit. Geometry-only CAD sketches are used to
evaluate linear decodability (P1/P2), matched forced-choice generation
(P3), activation-level influence, and steerability in frozen language
models. A limited-supervision follow-up tests whether LoRA can reduce
the P1--P3 gap relative to matched baselines.}
\label{fig:overview}
\end{figure}

\section{Related Work}

\subsection{Geometric Reasoning and CAD Generation}
Geometric constraint reasoning underlies engineering tasks such as 3D object placement, robotic assembly planning, and manufacturing workflows~\citep{huangFirePlaceGeometricRefinements2025,leuCADModelBased2013,gonzalez-lluchEffectsFixGeometric2019}. Parametric CAD makes this local-to-global structure explicit: pairwise geometric constraints define local relations, while remaining degrees of freedom characterize sketch-level constraint status. Prior work captures this structure through complementary representations. SketchGraphs models primitives and explicit pairwise constraints~\citep{seff2020sketchgraphs}, whereas Fusion~360 Gallery, DeepCAD, and CADParser represent designs through construction histories of increasing operational complexity~\citep{willisFusion360Gallery2021,wu2021deepcad,zhouCADParser2023}. Generative models accordingly target either constrained sketches, as in Vitruvion~\citep{seff2022vitruvion}, or CAD construction sequences, as in DeepCAD and SkexGen~\citep{wu2021deepcad,xu2022skexgen}. More recent systems such as Text2CAD, CAD-Llama, CADmium, STEP-LLM, and ReCAD leverage language models to generate CAD sequences or executable code~\citep{khan2024text2cad,liCADLlamaLeveragingLarge2025,govindarajanCADmiumFineTuningCode2025,shiSTEPLLMGeneratingCAD2026,liReCADReinforcementLearning2025}. Across these lines of work, evaluation has centered on task-level
outcomes such as conditional constraint prediction, CAD reconstruction,
validity, and generation quality. We instead ask whether frozen LLMs encode constraint structure, whether
that structure is expressed in generation, and whether interventions
on internal activations can influence or control predictions.

\subsection{Representation Analysis and Intervention}
A common approach to representation analysis is linear probing, which tests what information is linearly decodable from frozen hidden states~\citep{alain2017probing,belinkov2019analysis}. Beyond linguistic attributes, probing has identified linearly decodable world states in game-playing models~\citep{liEmergentWorldRepresentations2024,nandaEmergentLinearRepresentations2023} and spatial and temporal information in LLMs~\citep{gurneeLanguageModelsRepresent2024}. These findings are consistent with the linear representation hypothesis, which proposes that features are organized along directions in representation space~\citep{parkLinearRepresentationHypothesis2024}. Under this view, linear probing provides a natural tool for examining structured information in model representations.

However, high linear-probe accuracy does not by itself show that the relevant representation was learned through pretraining. Instead, it may reflect probe capacity, architectural bias, or information already present in the input~\citep{hewitt2019control,pimentel2020information}. Prior work addresses these alternatives using shuffled-label tasks and selectivity~\citep{hewitt2019control}, random encoders~\citep{wietingNoTrainingRequired2019}, and comparisons with simpler input representations~\citep{hewittConditionalProbingMeasuring2021}. Following these principles, we use shuffled-label and random-initialization controls and introduce a pure-input baseline for serialized geometry. 

Beyond identifying what probes can extract, prior work has developed interventions that test whether internal representations participate in model behavior~\citep{ravichanderProbingProbingParadigm2021}. Activation patching tests how interventions on intermediate states
affect model outputs~\citep{vig2020investigating,meng2022rome}.
Related intervention methods have been used to identify
behaviorally relevant circuits~\citep{wang2022interpretability,
conmy2023acdc}. Activation addition and representation engineering
instead modify internal states to steer model behavior
~\citep{turnerSteeringLanguageModels2024,
zouRepresentationEngineeringTopDown2025}. These methods test complementary aspects of representational function.
Patching asks whether restoring an activation at a selected position
affects the prediction, whereas steering asks whether modifying that
activation can control the output. We combine these interventions with
decoding and generation to distinguish linear accessibility,
behavioral expression, activation-level influence, and controllability.

\section{Method}
\label{sec:method}

We evaluate pretrained LLMs using geometry-only CAD serializations, three matched prediction tasks, and activation-level interventions. We then use LoRA with limited training data to test whether supervision
can recover accurate relation prediction when zero-shot generation
fails. Figure~\ref{fig:overview} summarizes the evaluation pipeline.

\subsection{Data and Matched Tasks}

We use the SketchGraphs training split~\citep{seff2020sketchgraphs}.
Each sketch is serialized as plain text containing entity types and
numeric parameters, with all \texttt{EdgeOp} annotations excluded.
Entity character spans are mapped to token spans for pooling. To
address class imbalance, we construct stratified, class-balanced
subsets of up to 15,000 pairs per P1 class and 12,000 sketches per P2
class, requiring at least 500 examples per class. The same sampled
pools and random seed are reused across models before model-specific
token-length filtering.

We define two decoding tasks and one matched generation task.
\textbf{P1} uses eight-class linear classification to decode the
relation of an ordered entity pair \((i,j)\) from
\(\mathbf{x}_{ij}=[\mathbf{h}_i;\mathbf{h}_j]\). The labels are
\textsc{Coincident}, \textsc{Parallel}, \textsc{Perpendicular},
\textsc{Tangent}, \textsc{Equal}, \textsc{Midpoint},
\textsc{Concentric}, and \textsc{NoConstraint}; the last is sampled
from unconstrained pairs in the same sketch. We exclude the primarily
unary \textsc{Horizontal} and \textsc{Vertical} relations.
\textbf{P2} uses three-class linear classification to decode
sketch-level degrees-of-freedom status from the mean-pooled entity
representation,
\[
\bar{\mathbf{h}}
=
|\mathcal{E}|^{-1}
\sum_{e\in\mathcal{E}}\mathbf{h}_e.
\]
Sketches are labeled as under-, well-, or over-constrained according
to whether the value returned by
\texttt{get\_sequence\_dof} is positive, zero, or negative. Because
this label is computed heuristically rather than verified by
solver-based rank analysis, redundant constraints may produce
incorrect P2 labels. \textbf{P3} is the zero-shot counterpart to P1:
given the same held-out pair, the frozen model completes
\mbox{\texttt{Constraint(E$i$,E$j$)} $= \ldots$}, and the predicted
class is selected using first-token forced-choice scores over the same
eight labels. The shared examples, label space, and macro-F1 metric
enable a matched comparison between P1 decodability and P3 generation.
\subsection{Probing, Controls, and Evaluation}

To measure where geometric information becomes linearly accessible
without changing the model weights, we extract hidden representations
from eight approximately evenly spaced relative-depth locations in
each model. At each location, token states within an entity span are
mean-pooled to obtain \(\mathbf{h}_e\), and an
\(\ell_2\)-regularized logistic regression probe is trained for each decoding
task. Relative-depth sampling enables comparison across architectures
with different numbers of layers. To evaluate generalization across
sketches, P1 uses a 75/25 sketch-grouped split that keeps all pairs
from the same sketch in one partition. P2 uses a class-stratified
75/25 split because each sketch contributes only one example.

We evaluate P1 and P2 using macro-F1, with uniform-class reference
levels of \(0.125\) and \(0.333\), respectively. Because high probe
performance may reflect learned representations, architectural
structure, raw input features, or probe capacity, we compare
pretrained models with several matched controls. Same-architecture random-init models provide a baseline for estimating
the contribution of pretraining.
Label-permutation probes provide the selectivity score
\[
\mathrm{Sel}(\ell)
=
F_{1,\mathrm{task}}(\ell)
-
F_{1,\mathrm{label\mbox{-}perm}}(\ell),
\]
which measures performance associated with the true labels beyond
that obtained from the same probe trained on permuted labels
\citep{hewitt2019control}. We also train a linear classifier, a
multilayer perceptron (MLP), and gradient-boosted trees (GBTs) on
numerical features derived directly from the entity geometry. The
nonlinear MLP and GBT baselines test whether the targets are
recoverable from the input without pretrained representations; these
input-only baselines do not enter DI. An entity-order shuffle tests
whether P1 depends on serialization order while preserving entity
identities, geometry, and labels, and an entity-count baseline tests
whether P2 can be explained by sketch size.

To compare the contribution of pretraining across the two decoding
tasks, we define a dissociation index (DI). For each architecture, we
first select the layer at which pretrained P1 performance is highest:
\begin{equation}
\label{eq:lstar}
\ell^*
=
\arg\max_{\ell}
F_{1,\mathrm{pre}}^{P1}(\ell).
\end{equation}
We then evaluate pretrained and random-init P1 and P2 at this same
layer:
\begin{equation}
\label{eq:di}
\mathrm{DI}
=
\left[
F_{1,\mathrm{pre}}^{P1}(\ell^*)
-
F_{1,\mathrm{rand}}^{P1}(\ell^*)
\right]
-
\left[
F_{1,\mathrm{pre}}^{P2}(\ell^*)
-
F_{1,\mathrm{rand}}^{P2}(\ell^*)
\right].
\end{equation}
A positive DI indicates a larger pretraining gain for local pairwise
relations than for global constraint status under the same-layer
comparison.
\subsection{Interventions and Limited-Supervision Adaptation}
We use activation patching on Qwen2.5-3B, Llama-3.1-8B, and
OLMo-3-7B to test whether entity representations affect P1 probe
predictions~\citep{vig2020investigating,meng2022rome,hanna2023gpt2greater}.
We add Gaussian noise with a standard deviation three times that of
the input hidden states to all input tokens corresponding to entity
\(i\), and retain only corruption-informative examples,
\[
\mathcal{I}
=
\{x:\hat{y}_{\mathrm{corr}}(x)
\neq \hat{y}_{\mathrm{clean}}(x)\}.
\]
At each layer \(\ell\), from the input through the model-specific P1
readout layer \(\ell^*\), we restore the clean activation of entity
\(i\) while leaving the remaining corrupted computation unchanged.
The restoration rate is
\[
\mathrm{RR}(\ell)
=
\frac{1}{|\mathcal{I}|}
\sum_{x\in\mathcal{I}}
\mathbf{1}
\!\left[
\hat{y}_{\mathrm{patch}}^{(\ell)}(x)
=
\hat{y}_{\mathrm{clean}}(x)
\right].
\]
Restoration therefore measures recovery of the clean prediction,
rather than macro-F1 or agreement with the ground-truth label.
Layer~0 directly reverses the input corruption and serves as a sanity
check. To measure spillover, we also record whether the post-patching
prediction for an unrelated pair \((k,m)\) from the same sketch
matches its clean prediction.

We separately test whether the representation space permits simple
targeted control of the model's first-token forced-choice predictions. For source
class \(a\) and target class \(b\), we define the normalized
mean-difference direction at layer \(\ell\) as
\[
\mathbf{d}_{a\rightarrow b}^{(\ell)}
=
\frac{\boldsymbol{\mu}_{b,\ell}-\boldsymbol{\mu}_{a,\ell}}
{\|\boldsymbol{\mu}_{b,\ell}-\boldsymbol{\mu}_{a,\ell}\|_2}.
\]
Here, \(\boldsymbol{\mu}_{c,\ell}\) is the class-conditional mean at
layer \(\ell\), computed from the average of the two queried entity
representations for P1 and the average over all entity representations
for P2. The class means are estimated from up to 1,000 examples per
class. We add
\(\alpha\mathbf{d}_{a\rightarrow b}^{(\ell)}\) at a common early
layer, \(\ell=4\), and at each model's readout layer \(\ell^*\). The
direction is applied to both queried entity spans for P1 and to all
entity spans for P2. We test
\textsc{NoConstraint}\(\rightarrow\)\textsc{Parallel} for P1 and
\textsc{UnderConstrained}\(\rightarrow\)\textsc{WellConstrained} for
P2 using \(\alpha\in\{0.5,1,2,4,8\}\) and 200 examples per task. We
report both target-class flips and unrestricted label changes, with
10 matched random directions as controls.

Finally, to test whether the P1--P3 gap can be reduced with limited
supervision, we fine-tune Qwen2.5-3B, Llama-3.1-8B, and OLMo-3-7B on
P3 using LoRA. Each training set contains 50 sketch-disjoint pairs per
class, or 400 pairs in total, and training is repeated across three
sampling and initialization seeds. We compare LoRA with frozen linear
probes and input-only GBTs trained on the same examples. The GBTs use
either raw geometric features or the numerical values appearing in
the serialized prompt. For each model, all methods are evaluated on the same balanced set of
2,000 held-out P3 pairs.

\subsection{Experimental Setup}

We evaluate eight pretrained language models spanning four model
families and multiple scales: Qwen2.5-0.5B, 1.5B, 3B, and
7B~\citep{qwenQwen25TechnicalReport2025}, Qwen3.5-9B-Base~\citep{qwen3.5},
Mistral-7B~\citep{jiang2023mistral},
Llama-3.1-8B~\citep{grattafiori2024llama3herdmodels}, and OLMo-3-7B~\citep{olmoOlmo32026}.
Each model is paired with a randomly initialized counterpart of the
same architecture, and all models use the same pre-sampled sketch
pools. Because tokenization and length filtering produce slightly
different retained subsets, we also evaluate DI on the intersection
of sketches retained by all eight models.

The main probe split and zero-shot P3 sample use seed~0. P3 selects
the label with the highest uncalibrated first-token score, while
content-free calibration is reported separately in
Appendix~\ref{app:p3_diagnostics}. Reported 95\% confidence intervals
are estimated using 1,000 bootstrap resamples. DI uncertainty is
obtained by combining the P1 and P2 component uncertainties in
quadrature because the tasks use different evaluation units. For the
limited-supervision experiments, LoRA is applied to all linear modules
with rank~16 and trained for 10 epochs using completion-only
cross-entropy over the target-label tokens.%
\footnote{LoRA uses scaling factor~32 and dropout~0.05. Optimization
uses learning rate \(2\times10^{-4}\), effective batch size~16,
maximum sequence length~1024, and bfloat16 precision.}
LoRA and the matched low-data controls are repeated across three data
sampling and initialization seeds and reported as mean and sample
standard deviation.
\section{Results}
\subsection{Constraint Information Is Linearly Decodable}
\label{sec:r1}

\begin{figure*}[t]
\centering
\includegraphics[width=\linewidth]{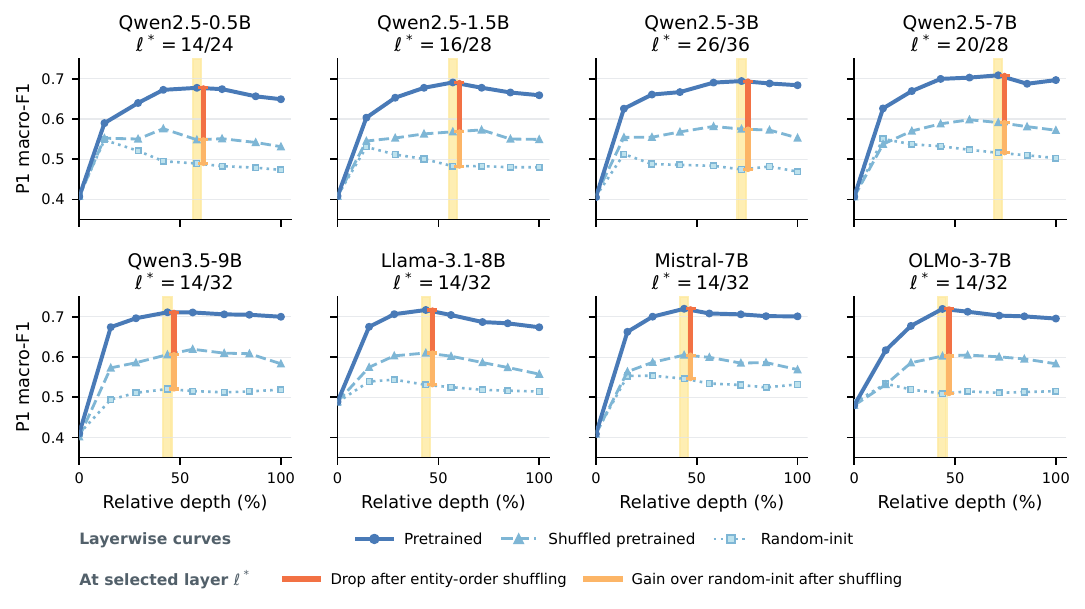}
\caption{P1 macro-F1 across relative depth for eight models. At each pretrained P1 peak \(\ell^*\), the vertical segments decompose the pretrained–random difference into the drop after entity-order shuffling and the remaining shuffled-pretrained advantage over random initialization.}
\label{fig:p1_curves}
\end{figure*}

Both P1 and P2 are linearly decodable from the hidden states of all eight
pretrained models. Peak P1 macro-F1 ranges from 0.678 to 0.720, well above
the 0.125 chance level (Table~\ref{tab:multimodel}). Selectivity remains high at 0.558--0.597, indicating that the strong
P1 performance is not explained by fitting shuffled labels. Across architectures, P1 decodability rises rapidly in early layers
and remains high across a broad depth range
(Figure~\ref{fig:p1_curves}). At the P1-selected layers, P2 macro-F1
is similarly high at 0.718--0.739, although Section~\ref{sec:r2} shows
that random-init representations already reach 0.676--0.690.

\subsection{P1 Benefits More from Pretraining Than P2}
\label{sec:r2}

For P1, pretraining improves performance over random initialization
by 0.174--0.220. A linear classifier on the raw parameters reaches
0.372, while a gradient-boosted tree (GBT) reaches approximately 0.70,
showing that the relations are recoverable from the input and that
pretraining makes comparable information accessible to a linear
probe. Shuffling entity order lowers pretrained P1 by 0.107--0.129.
Nevertheless, the shuffled pretrained point estimates remain
0.059--0.100 above the corresponding canonical-order random-init
baselines (Figure~\ref{fig:p1_curves}). Thus, entity order provides a
substantial cue but does not fully explain the P1 advantage.

P2 relies much less on pretraining. At the P1-selected layer
\(\ell^*\), pretraining improves P2 over random initialization by only
0.041--0.051, compared with 0.174--0.220 for P1
(Table~\ref{tab:multimodel}). Entity count alone reaches 0.419, and a
GBT on the raw parameters reaches approximately 0.70, close to the
random-init P2 performance of 0.676--0.690. Thus, most of P2's
decodability is available without pretrained weights. Consistent with
this asymmetry, DI is positive for all eight models (0.128--0.169),
with every 95\% confidence interval excluding zero. Overall,
pretraining contributes substantially more to P1 decodability than
to P2.
\begin{table*}[t]
\centering
\small
\setlength{\tabcolsep}{4pt}
\begin{tabular}{@{}lccccccc@{}}
\toprule
& \multicolumn{2}{c}{Macro-F1}
& & \multicolumn{2}{c}{Pretraining gains}
& \multicolumn{1}{c}{Primary contrast}
& \multicolumn{1}{c}{Order control} \\
\cmidrule(lr){2-3}
\cmidrule(lr){5-6}
\cmidrule(lr){7-7}
\cmidrule(l){8-8}
Model
& P1
& P2
& $\ell^*/L$
& $\Delta_{\mathrm{pre}}^{P1}$
& $\Delta_{\mathrm{pre}}^{P2}$
& DI [95\% CI]
& Shuf.$-$Rand. \\
\midrule

\multicolumn{8}{l}{
\textbf{Input-only baselines}
\quad\textit{(Llama-retained split)}
} \\
Chance
  & .125 & .333 & -- & -- & -- & -- & -- \\
Entity count
  & -- & .419 & -- & -- & -- & -- & -- \\
Linear, raw parameters
  & .372 & .384 & -- & -- & -- & -- & -- \\
MLP, raw parameters
  & .560 & .612 & -- & -- & -- & -- & -- \\
GBT, raw parameters
  & .697 & .695 & -- & -- & -- & -- & -- \\

\midrule
\multicolumn{8}{l}{\textbf{Random-init representations}} \\
Qwen2.5-0.5B
  & .489 & .676 & 14/24 & -- & -- & -- & -- \\
Qwen2.5-1.5B
  & .483 & .688 & 16/28 & -- & -- & -- & -- \\
Qwen2.5-3B
  & .475 & .688 & 26/36 & -- & -- & -- & -- \\
Qwen2.5-7B
  & .516 & .686 & 20/28 & -- & -- & -- & -- \\
Qwen3.5-9B
  & .521 & .681 & 14/32 & -- & -- & -- & -- \\
Llama-3.1-8B
  & .531 & .681 & 14/32 & -- & -- & -- & -- \\
Mistral-7B
  & .546 & .682 & 14/32 & -- & -- & -- & -- \\
OLMo-3-7B
  & .509 & .690 & 14/32 & -- & -- & -- & -- \\

\midrule
\multicolumn{8}{l}{\textbf{Pretrained representations}} \\
Qwen2.5-0.5B
  & .678 & .718 & 14/24
  & .188 & .043
  & .145 [.120, .171] & .059 \\
Qwen2.5-1.5B
  & .691 & .732 & 16/28
  & .208 & .045
  & .164 [.139, .188] & .086 \\
Qwen2.5-3B
  & .694 & .739 & 26/36
  & .220 & .051
  & .169 [.144, .193] & .100 \\
Qwen2.5-7B
  & .709 & .727 & 20/28
  & .193 & .041
  & .152 [.127, .176] & .075 \\
Qwen3.5-9B
  & .711 & .729 & 14/32
  & .191 & .048
  & .143 [.119, .166] & .086 \\
Llama-3.1-8B
  & .717 & .728 & 14/32
  & .186 & .047
  & .140 [.115, .164] & .079 \\
Mistral-7B
  & .720 & .728 & 14/32
  & .174 & .046
  & .128 [.104, .153] & .059 \\
OLMo-3-7B
  & .720 & .731 & 14/32
  & .211 & .042
  & .169 [.145, .192] & .093 \\
\bottomrule
\end{tabular}

\caption{
Main representation results at each architecture's pretrained
P1-selected layer $\ell^*$. Random-init representations are evaluated
at the same layer. The pretraining gain is
$\Delta_{\mathrm{pre}}^{t}
=F_{1,\mathrm{pre}}^{t}(\ell^*)
-F_{1,\mathrm{rand}}^{t}(\ell^*)$.
DI is the primary contrast between the P1 and P2 gains.
Shuf.$-$Rand.\ is the residual pretrained P1 advantage after
entity-order shuffling. Input-only baselines use no LLM
representations and are evaluated on the Llama-retained split.
}
\label{tab:multimodel}
\end{table*}
\subsection{The Dissociation Holds Across Scale and Architecture}
\label{sec:r3}

We tested whether the P1--P2 dissociation extends beyond a single
model. Within the Qwen2.5 family, P1 macro-F1 increases monotonically
from 0.678 to 0.709 as model size grows, whereas P2 ranges from 0.718
to 0.739 and peaks at 3B (Table~\ref{tab:multimodel}). DI remains
positive at every scale, ranging from 0.145 to 0.169, but varies
non-monotonically. Scaling therefore modestly improves absolute P1
decodability without consistently increasing P2 or the relative
pretraining advantage measured by DI.

The dissociation also holds across model families. Among the four
approximately size-matched 7--8B models, P1 macro-F1 ranges from 0.709
to 0.720, P2 from 0.727 to 0.731, and DI from 0.128 to 0.169, with
every 95\% confidence interval excluding zero. The selected P1 peaks
span layers 14 to 26, or approximately 44\% to 72\% of model depth,
indicating that the pattern is not tied to a shared absolute or
relative layer. Chance-normalized
\(\mathrm{DI}_{\mathrm{norm}}\) also remains positive across all
models (0.130--0.178; Table~\ref{tab:normalized-di},
Appendix~\ref{app:di_norm}). Restricting every model to the common set
of retained sketches yields similar DI values of 0.129--0.177, all
within the corresponding full-set confidence intervals. Together,
these results show that the dissociation persists across the tested
scales, architectures, chance normalization, and retained-set
differences.
\subsection{Generation Falls Short of Decodability}
\label{sec:p3}

For each model, we compare P3 forced-choice generation with P1
decoding on the same model-specific, class-balanced sample of 2,000
held-out pairs. The samples may differ across models, but each contains
250 examples from each of the eight constraint classes. Across all
eight models, P3 macro-F1 ranges from 0.028 to 0.344, substantially
below the matched P1 range of 0.683 to 0.725 and yielding P1--P3 gaps
of 0.381 to 0.686 (Table~\ref{tab:p1_p3}). Thus, linearly decodable
constraint information is not reliably expressed in the models' own
predictions.

\begin{table}[t]
\caption{Matched P1 probe and zero-shot P3 generation macro-F1.
For each model, both are evaluated on the same model-specific,
class-balanced sample of 2,000 held-out pairs
(250 per class; chance${}=0.125$). P1 is evaluated at \(\ell^*\).
Gap${}={}$P1${}-$P3.}
\label{tab:p1_p3}
\centering
\begin{tabular}{lccc}
\toprule
Model & P1 probe F1 & P3 gen.\ F1 & Gap \\
\midrule
Qwen2.5-0.5B & 0.683 & 0.095 & 0.588 \\
Qwen2.5-1.5B & 0.699 & 0.072 & 0.628 \\
Qwen2.5-3B   & 0.691 & 0.089 & 0.602 \\
Qwen2.5-7B   & 0.709 & 0.257 & 0.451 \\
Qwen3.5-9B   & 0.725 & 0.344 & 0.381 \\
Mistral-7B   & 0.714 & 0.028 & 0.686 \\
Llama-3.1-8B & 0.716 & 0.141 & 0.575 \\
OLMo-3-7B    & 0.716 & 0.129 & 0.587 \\
\bottomrule
\end{tabular}
\end{table}

\begin{figure}[t]
\centering
\includegraphics[width=0.49\linewidth]
{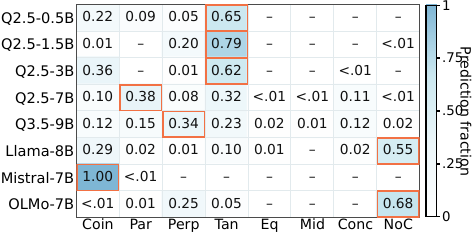}
\caption{P3 predicted-class distributions across eight models.
Rows denote models and columns denote predicted classes. Boxes mark
each model's dominant predicted class; dashes denote exact zeros.
Per-class accuracies are reported in
Table~\ref{tab:p3-per-class}.}
\label{fig:p3_matrix}
\end{figure}

Failure modes differ across architectures
(Figure~\ref{fig:p3_matrix}). Mistral-7B assigns 99.8\% of its
predictions to \texttt{Coincident}, producing an almost complete
single-class collapse, while OLMo-3-7B assigns 68.4\% to
\texttt{NoConstraint}. Qwen3.5-9B achieves the highest P3 macro-F1 of
0.344, followed by Qwen2.5-7B at 0.257. Content-free calibration, which addresses architecture-specific
output-label preferences, improves P3 macro-F1 from 0.028--0.344 to
0.142--0.380, but the gap from the matched P1 probes remains
0.345--0.562 (Table~\ref{tab:p3-calibration}). On Qwen2.5-3B,
four-shot prompting increases macro-F1 from 0.089 to
\(0.138 \pm 0.013\) across five exemplar sets, but remains well below
the matched P1 score. Thus, output-label preferences and limited
in-context examples account for only part of the generation gap.
Full per-class results are reported in
Table~\ref{tab:p3-per-class}.

\subsection{Activation-Level Influence and Steerability}
\label{sec:r4}

\begin{figure*}[t]
    \centering
    \includegraphics[width=\textwidth]
        {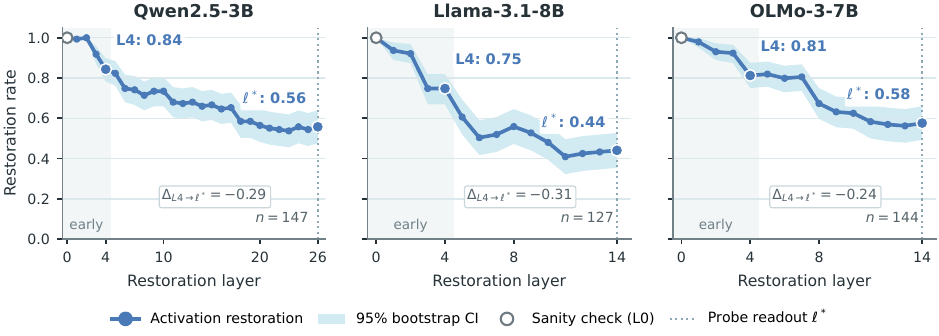}
\caption{Layerwise P1 restoration rates across three models.
Restoration generally declines with depth but remains substantial at
the probe readout $\ell^*$. Shading denotes 95\% bootstrap confidence
intervals.}
    \label{fig:patch_grid}
\end{figure*}

We first test whether entity representations causally affect the P1
probe prediction. For each model, we corrupt the input embeddings of
entity \(i\) and retain only examples for which this intervention
changes the clean prediction, yielding 147 informative examples for
Qwen2.5-3B, 127 for Llama-3.1-8B, and 144 for OLMo-3-7B. We then
restore the clean representation of \(i\) at each layer and evaluate
the frozen probe at the model's P1 peak layer \(\ell^*\). Recovery is
nearly complete when restoration occurs in the first one or two
layers. At layer~4, restoration rates range from 0.748 to 0.844; they
generally decline with depth but remain substantial at the probe
readout, ranging from 0.441 to 0.576
(Figure~\ref{fig:patch_grid}). Thus, the effect of the corrupted entity
is not confined to an early, transient stage. Incomplete recovery at the readout may reflect the propagation of
corruption to the paired entity $j$. The unrelated-pair
control further shows that the intervention propagates beyond the
directly corrupted entity: corrupting \(i\) changes the clean
prediction for an unrelated pair in 34.6--35.4\% of informative
examples, while restoration at layer~4 returns that prediction to its
clean value in 76.4--84.7\% of examples.

We test targeted mean-difference steering at layer~4 and the P1 peak
layer (\(26\) for Qwen and \(14\) for Llama and OLMo), using
\textsc{NoConstraint}\(\rightarrow\)\textsc{Parallel} for P1 and
\textsc{UnderConstrained}\(\rightarrow\)\textsc{WellConstrained} for
P2. Under first-token scoring, P1 target-class flips remain rare: at
\(\alpha=8\), rates range from 0 to 0.070 and do not consistently
exceed matched random-direction controls (0.005--0.082). For P2, no
target-class flips occur for any tested model, layer, or steering
strength.
This contrasts with patching: restoring entity activations can recover
frozen-probe predictions, but moving activations along a global
mean-difference direction does not reliably control the model's
output. Together, these results show that linear decodability and
causal influence do not, by themselves, imply linear steerability.
\subsection{Limited Supervision Recovers Relation Prediction}
\label{sec:lora}

The P1--P3 gap shows that relations can be decoded from hidden states
even when zero-shot generation fails to express them. We therefore
test whether limited supervision can reduce this gap and whether
comparable performance can be learned directly from numerical CAD
inputs. LoRA raises P3 macro-F1 from the zero-shot range of 0.089--0.141
to 0.637--0.699 (Table~\ref{tab:lora}). With the same 400-example
training budget, frozen probes reach 0.611--0.627, whereas input-only
GBTs reach at most 0.484. Scoring the complete label sequence instead
of only its first token changes LoRA macro-F1 by less than 0.001.
Llama and OLMo use the same input-only split, which yields identical
GBT results.

\begin{table}[t]
\caption{First-token forced-choice macro-F1 on matched P3 test pairs.
Supervised methods use 50 sketch-disjoint training pairs per class
(400 total). Values are mean $\pm$ sample standard deviation over
three seeds. Zero-shot P3 is evaluated once.}
\label{tab:lora}

\begin{center}
\setlength{\tabcolsep}{4pt}
\begin{tabular}{@{}lccccc@{}}
\toprule
& Zero-shot
& \multicolumn{2}{c}{LLM-based supervision}
& \multicolumn{2}{c}{Input-only supervision} \\
\cmidrule(lr){3-4}\cmidrule(lr){5-6}
Model & P3 & LoRA SFT & Frozen probe & Raw GBT & Prompt GBT \\
\midrule
Qwen2.5-3B & .089 & \textbf{.637$\pm$.008} & .611$\pm$.006
& .484$\pm$.022 & .446$\pm$.008 \\
Llama-3.1-8B & .141 & \textbf{.680$\pm$.007} & .627$\pm$.022
& .475$\pm$.018 & .441$\pm$.016 \\
OLMo-3-7B & .129 & \textbf{.699$\pm$.015} & .624$\pm$.021
& .475$\pm$.018 & .441$\pm$.016 \\
\bottomrule
\end{tabular}
\end{center}
\end{table}

The advantage of frozen probes over input-only GBTs indicates that
pretrained hidden states support more sample-efficient relation
prediction in this low-data setting. LoRA further makes accurate
relation prediction available through the model's output, contrasting
with the weak and inconsistent effects of mean-difference steering in
Section~\ref{sec:r4}. However, LoRA can modify internal
representations as well as their mapping to output labels, so its
success does not show that fine-tuning merely redirects an unchanged
representation. Nor does the low-data advantage imply that the
relation is absent from the CAD input: with larger training sets,
raw-feature GBTs approach 0.70 macro-F1. The advantage observed here
therefore concerns low-data sample efficiency rather than exclusive
access to the relation.
% St

\section{Discussion}

\subsection{Possible Sources of the Local--Global Pretraining Asymmetry}

The consistently positive DI indicates that pretraining contributes
more to local relation decoding than to global constraint-status
decoding. One possible explanation is that P1 requires representations
of relative geometry that generalize across particular coordinates and
entity order. The remaining pretrained-over-random gain after
entity-order shuffling is consistent with this interpretation. In contrast, P2 is highly decodable even from random-init representations, likely because its labels are recoverable from a combination of input-level features rather than entity count alone (\(0.419\) macro-F1), as supported by raw-feature GBTs approaching \(0.70\). A same-architecture random-init model may similarly transform these features into representations that a linear probe can exploit, consistent with prior work motivating random-representation baselines and probing controls~\citep{zhangLanguageModelingTeaches2018,hewitt2019control}. Thus, pretraining may still contribute to P2, but this contribution is better reflected by the pretrained-over-random gain than by absolute probe performance.

\subsection{The Gap Between Linear Access and Generation}
The P1--P3 gap shows that relation information accessible to a
supervised linear probe is not necessarily expressed through the
frozen model's generation pathway. 
Prior work shows that probe decodability need not predict
behavioral use~\citep{ravichanderProbingProbingParadigm2021, elazar2021amnesic}. One possible explanation for the P1--P3 gap is that the model prefers
particular label tokens regardless of the geometric input. We test this
possibility using content-free calibration
\citep{zhaoCalibrateUseImproving2021}. Calibration narrows but does not
close the gap, indicating that label-token preferences provide only a
partial explanation. We next ask whether limited supervision can reduce the P1--P3 gap. LoRA substantially improves P3, and both LoRA and frozen
probes outperform input-only GBTs under the same low-data budget. This
is consistent with pretrained representations providing useful
relational features. However, because LoRA can modify both the
representations and their readout, its success does not mean that the zero-shot model merely lacked the right output mapping.

\subsection{Intervention Scope and Access to Relational Information}

The contrast among patching, steering, and LoRA suggests that relational information is not equally accessible through different interventions. Patching restores the entire example-specific activation and can repair corrupted probe predictions. Steering instead modifies only one shared direction across examples, which may explain why it does not reliably produce target-class flips. LoRA makes broader
changes across multiple modules and substantially improves generation,
but this does not reveal whether it accesses information already
present or learns new relational features. Because patching is evaluated through frozen-probe predictions,
whereas steering and LoRA are evaluated through model outputs, these
results provide a qualitative rather than direct comparison.

Intervention timing may explain why restoration is strongest in early
layers. Patching an earlier layer leaves more subsequent layers to use the restored activation, so stronger early-layer recovery does not necessarily mean that relational information exists only in early layers. Together, early patching recovery and unreliable steering suggest that the model may rely on multiple context-dependent features rather than a single shared direction, although identifying the mechanism requires circuit-level analysis.

\subsection{Limitations}
\label{sec:limitations}

Our conclusions are limited by the scope of the data and labels.
The primary analysis uses SketchGraphs, while the Fusion~360
cross-dataset evaluation covers only P1. We therefore cannot establish
that the full P1--P2 dissociation generalizes across CAD datasets. P1
also excludes the primarily unary \texttt{Horizontal} and
\texttt{Vertical} relations. P2 relies on the heuristic
\texttt{get\_sequence\_dof} count, which does not verify constraint
independence and may mislabel redundant constraint sets. Its conclusions are therefore limited to coarse DOF status rather than whether the constraints are consistent, redundant, or admit a valid solution.

Additionally, our experiments do not capture all sources of
variation. The main P1 and P2 analyses use one data split and one
random-init checkpoint per architecture, while the entity-order and P3
evaluations use fixed sampling seeds. Our confidence intervals capture
variation across test examples, but not across data splits, model
initializations, or peak-layer selection, and may therefore understate
the total uncertainty.

Finally, the behavioral and intervention results cover fewer settings
than the decoding analysis. P3 uses one forced-choice format, while
patching, steering, and LoRA are evaluated on three models. Patching
measures frozen-probe recovery only on corruption-informative examples,
and steering tests two class transitions using mean-difference
directions, so the conclusions are limited to these specific
interventions. 

\subsection{Future Work}
\label{sec:future}

Future work can test whether the local--global pretraining asymmetry
extends beyond SketchGraphs and general-purpose language models. A
stronger cross-dataset test would extend the analysis to P2 using
reliable global constraint labels, ideally verified by a solver.
Evaluating CAD-native models such as
Vitruvion~\citep{seff2022vitruvion} could show whether domain-specific
training changes the observed dissociation. Mechanistic studies could
track how restored activations propagate across token positions,
layers, and model components, and test steering methods that use
multiple or example-specific directions. Comparing representations
before and after LoRA could also help determine whether fine-tuning
teaches the model to report information it already had or changes how
the relation is represented.
\section{Conclusion}

In this study, we introduced a controlled framework for testing whether geometric
information in frozen language models is encoded, expressed in their
predictions, and accessible to activation-level control. Across eight
pretrained models and matched random-init controls, pretraining
contributes substantially more to decoding local pairwise relations
than to classifying sketch-level degrees-of-freedom status. Because
global constraint status is already highly decodable from random-init
representations, high probe performance alone cannot be attributed to
pretraining.

Local relation information also exhibits a clear
representation--behavior gap: zero-shot generation performs
substantially below matched linear probes. On three representative
models, activation patching repairs corrupted probe predictions,
whereas mean-difference steering does not provide reliable targeted
control. Limited LoRA supervision, however, substantially improves
relation prediction. Together, these findings show that information
can be linearly decoded and influence probe predictions without being
reliably expressed or controlled in model behavior. For practical CAD
systems, internal decodability should therefore not substitute for
output-level evaluation: generated constraints should be checked
directly and, when appropriate, verified with a solver.

\bibliography{main}

@inproceedings{seff2020sketchgraphs,
  author    = {Seff, Ari and Ovadia, Yaniv and Zhou, Wenda and Adams, Ryan P.},
  title     = {SketchGraphs: A Large-Scale Dataset for Modeling Relational Geometry in {CAD}},
  booktitle = {ICML Workshop on Object-Oriented Learning},
  year      = {2020}
}

@inproceedings{seff2022vitruvion,
  author    = {Seff, Ari and Zhou, Wenda and Richardson, Nathan and Adams, Ryan P.},
  title     = {Vitruvion: A Generative Model of Parametric {CAD} Sketches},
  booktitle = {International Conference on Learning Representations},
  year      = {2022}
}

@inproceedings{wu2021deepcad,
  author    = {Wu, Rundi and Xiao, Chang and Zheng, Changxi},
  title     = {Deep{CAD}: A Deep Generative Network for Computer-Aided Design Models},
  booktitle = {Proceedings of the IEEE/CVF International Conference on Computer Vision (ICCV)},
  year      = {2021},
  pages     = {6772--6782}
}

@inproceedings{xu2022skexgen,
  author    = {Xu, Xiang and Willis, Karl D. D. and Lambourne, Joseph G. and Cheng, Chin-Yi and Jayaraman, Pradeep Kumar and Furukawa, Yasutaka},
  title     = {Skex{G}en: Autoregressive Generation of {CAD} Construction Sequences with Disentangled Codebooks},
  booktitle = {International Conference on Machine Learning},
  year      = {2022},
  pages     = {24698--24724}
}

@inproceedings{khan2024text2cad,
  author    = {Khan, Mohammad Sadil and Sinha, Sankalp and Sheikh, Talha Uddin and Stricker, Didier and Ali, Sk Aziz and Afzal, Muhammad Zeshan},
  title     = {Text2{CAD}: Generating Sequential {CAD} Designs from Beginner-to-Expert Level Text Prompts},
  booktitle = {Advances in Neural Information Processing Systems},
  year      = {2024},
  pages     = {7552--7579}
}

@misc{grattafiori2024llama3herdmodels,
      title={The Llama 3 Herd of Models}, 
      author={Aaron Grattafiori and Abhimanyu Dubey and Abhinav Jauhri and Abhinav Pandey and Abhishek Kadian and others},
      year={2024},
      eprint={2407.21783},
      archivePrefix={arXiv},
      primaryClass={cs.AI},
      url={https://arxiv.org/abs/2407.21783}, 
}

@inproceedings{alain2017probing,
  author    = {Alain, Guillaume and Bengio, Yoshua},
  title     = {Understanding Intermediate Layers Using Linear Classifier Probes},
  booktitle = {ICLR Workshop},
  year      = {2017}
}

@inproceedings{hewitt2019control,
  author    = {Hewitt, John and Liang, Percy},
  title     = {Designing and Interpreting Probes with Control Tasks},
  booktitle = {Empirical Methods in Natural Language Processing},
  year      = {2019}
}

@article{belinkov2019analysis,
  author    = {Belinkov, Yonatan and Glass, James},
  title     = {Analysis Methods in Neural Language Processing: A Survey},
  journal   = {Transactions of the Association for Computational Linguistics},
  volume    = {7},
  pages     = {49--72},
  year      = {2019}
}

@inproceedings{pimentel2020information,
  author    = {Pimentel, Tiago and Pimentel, Josef and Velioglu, Huseyin and Wich, Max and Cotterell, Ryan},
  title     = {Information-Theoretic Probing for Linguistic Structure},
  booktitle = {Association for Computational Linguistics},
  year      = {2020},
  pages     = {4609--4619}
}

@inproceedings{elazar2021amnesic,
  author    = {Elazar, Yanai and Kamfonas, Michael and Ravfogel, Shauli and Goldberg, Yoav},
  title     = {Amnesic Probing: Behavioral Explanation with Amnesic Counterfactuals},
  booktitle = {Association for Computational Linguistics},
  year      = {2021},
  pages     = {2764--2773}
}

@inproceedings{vig2020investigating,
  author    = {Vig, Jesse and Gehrmann, Sebastian and Belinkov, Yonatan and Qian, Sophie and Nevo, Daniel and Singer, Yaron and Shieber, Stuart},
  title     = {Investigating Gender Bias in Language Models Using Causal Mediation Analysis},
  booktitle = {Proceedings of the 58th Annual Meeting of the Association for Computational Linguistics},
  year      = {2020},
  pages     = {4548--4561}
}

@inproceedings{meng2022rome,
  author    = {Meng, Kevin and Bau, David and Andonian, Alex and Belinkov, Yonatan},
  title     = {Locating and Editing Factual Associations in {GPT}},
  booktitle = {Advances in Neural Information Processing Systems},
  year      = {2022},
  volume    = {35}
}

@inproceedings{hanna2023gpt2greater,
  author    = {Hanna, Michael and Liu, Ollie and Variengien, Alexandre},
  title     = {How Does {GPT}-2 Compute Greater-than?: Interpreting Mathematical Abilities in a Pre-Trained Language Model},
  booktitle = {Advances in Neural Information Processing Systems},
  year      = {2023}
}

@inproceedings{conmy2023acdc,
  author    = {Conmy, Arthur and Mavor-Parker, August and Lynch, Aengus and Heimersheim, Stef and Garriga-Alonso, Adri{\`a}},
  title     = {Towards Automated Circuit Discovery for Mechanistic Interpretability},
  booktitle = {Advances in Neural Information Processing Systems},
  year      = {2023}
}

@inproceedings{wang2022interpretability,
  author    = {Wang, Kevin and Variengien, Alexandre and Conmy, Arthur and Shlegeris, Buck and Steinhardt, Jacob},
  title     = {Interpretability in the Wild: A Circuit for Indirect Object Identification in {GPT}-2 Small},
  booktitle = {International Conference on Learning Representations},
  year      = {2023}
}

@article{jiang2023mistral,
  author    = {Jiang, Albert Q. and Sablayrolles, Alexandre and Mensch, Arthur and Bamford, Chris and Chaplot, Devendra Singh and Casas, Diego de las and Bressand, Florian and Lengyel, Gianna and Lample, Guillaume and Saulnier, Lucile and others},
  title     = {Mistral 7B},
  journal   = {arXiv preprint arXiv:2310.06825},
  year      = {2023}
}

@article{huangFirePlaceGeometricRefinements2025,
    address = {Nashville, TN, USA},
    title = {{FirePlace}: {Geometric} {Refinements} of {LLM} {Common} {Sense} {Reasoning} for {3D} {Object} {Placement}},
    copyright = {https://doi.org/10.15223/policy-029},
    shorttitle = {{FirePlace}},
    url = {https://ieeexplore.ieee.org/document/11094419/},
    doi = {10.1109/CVPR52734.2025.01257},
    urldate = {2026-08-13},
    journal = {2025 IEEE/CVF Conference on Computer Vision and Pattern Recognition (CVPR)},
    publisher = {IEEE},
    author = {Huang, Ian and Bao, Yanan and Truong, Karen and Zhou, Howard and Schmid, Cordelia and Guibas, Leonidas and Fathi, Alireza},
    month = jun,
    year = {2025},
    pages = {13466--13476},
}

@article{gonzalez-lluchEffectsFixGeometric2019,
    title = {On the effects of the fix geometric constraint in {2D} profiles on the reusability of parametric {3D} {CAD} models},
    volume = {29},
    issn = {0957-7572, 1573-1804},
    url = {http://link.springer.com/10.1007/s10798-018-9458-z},
    doi = {10.1007/s10798-018-9458-z},
    language = {en},
    number = {4},
    urldate = {2026-08-13},
    journal = {International Journal of Technology and Design Education},
    author = {González-Lluch, Carmen and Company, Pedro and Contero, Manuel and Pérez-López, David and Camba, Jorge D.},
    month = sep,
    year = {2019},
    pages = {821--841},
}

@article{leuCADModelBased2013,
    title = {{CAD} model based virtual assembly simulation, planning and training},
    volume = {62},
    copyright = {https://www.elsevier.com/tdm/userlicense/1.0/},
    issn = {00078506},
    url = {https://linkinghub.elsevier.com/retrieve/pii/S0007850613001959},
    doi = {10.1016/j.cirp.2013.05.005},
    language = {en},
    number = {2},
    urldate = {2026-08-13},
    journal = {CIRP Annals},
    author = {Leu, Ming C. and ElMaraghy, Hoda A. and Nee, Andrew Y.C. and Ong, Soh Khim and Lanzetta, Michele and Putz, Matthias and Zhu, Wenjuan and Bernard, Alain},
    year = {2013},
    pages = {799--822},
}

@incollection{zhouCADParser2023,
    series = {Guide {Proceedings}},
    title = {{CADParser}},
    url = {https://dl.acm.org/doi/10.24963/ijcai.2023/200},
    doi = {10.24963/ijcai.2023/200},
    urldate = {2026-08-13},
    booktitle = {Proceedings of the {Thirty}-{Second} {International} {Joint} {Conference} on {Artificial} {Intelligence}},
    author = {Zhou, Shengdi and Tang, Tianyi and Zhou, Bin},
    month = aug,
    year = {2023},
    pages = {1804--1812},
}

@article{govindarajanCADmiumFineTuningCode2025,
    title = {{CADmium}: {Fine}-{Tuning} {Code} {Language} {Models} for {Text}-{Driven} {Sequential} {CAD} {Design}},
    copyright = {Creative Commons Attribution 4.0 International},
    shorttitle = {{CADmium}},
    url = {https://arxiv.org/abs/2507.09792},
    doi = {10.48550/ARXIV.2507.09792},
    urldate = {2026-08-13},
    publisher = {arXiv},
    author = {Govindarajan, Prashant and Baldelli, Davide and Pathak, Jay and Fournier, Quentin and Chandar, Sarath},
    year = {2025},
}

@article{shiSTEPLLMGeneratingCAD2026,
    title = {{STEP}-{LLM}: {Generating} {CAD} {STEP} {Models} from {Natural} {Language} with {Large} {Language} {Models}},
    copyright = {arXiv.org perpetual, non-exclusive license},
    shorttitle = {{STEP}-{LLM}},
    url = {https://arxiv.org/abs/2601.12641},
    doi = {10.48550/ARXIV.2601.12641},
    urldate = {2026-08-13},
    publisher = {arXiv},
    author = {Shi, Xiangyu and Ding, Junyang and Zhao, Xu and Zhan, Sinong and Mohapatra, Payal and Quispe, Daniel and Welbeck, Kojo and Cao, Jian and Chen, Wei and Guo, Ping and Zhu, Qi},
    year = {2026},
}

@article{liCADLlamaLeveragingLarge2025,
    address = {Nashville, TN, USA},
    title = {{CAD}-{Llama}: {Leveraging} {Large} {Language} {Models} for {Computer}-{Aided} {Design} {Parametric} {3D} {Model} {Generation}},
    copyright = {https://doi.org/10.15223/policy-029},
    shorttitle = {{CAD}-{Llama}},
    url = {https://ieeexplore.ieee.org/document/11094068/},
    doi = {10.1109/CVPR52734.2025.01730},
    urldate = {2026-08-13},
    journal = {2025 IEEE/CVF Conference on Computer Vision and Pattern Recognition (CVPR)},
    publisher = {IEEE},
    author = {Li, Jiahao and Ma, Weijian and Li, Xueyang and Lou, Yunzhong and Zhou, Guichun and Zhou, Xiangdong},
    month = jun,
    year = {2025},
    pages = {18563--18573},
}

@inproceedings{liReCADReinforcementLearning2025,
    title = {{ReCAD}: {Reinforcement} {Learning} {Enhanced} {Parametric} {CAD} {Model} {Generation} with {Vision}-{Language} {Models}},
    copyright = {Creative Commons Attribution 4.0 International},
    shorttitle = {{ReCAD}},
    url = {https://arxiv.org/abs/2512.06328},
    doi = {10.48550/ARXIV.2512.06328},
    urldate = {2026-08-13},
    publisher = {arXiv},
    author = {Li, Jiahao and Luo, Yusheng and Lou, Yunzhong and Zhou, Xiangdong},
    year = {2025},
}

@inproceedings{wuAutoGenEnablingNextGen2023,
    title = {{AutoGen}: {Enabling} {Next}-{Gen} {LLM} {Applications} via {Multi}-{Agent} {Conversation}},
    booktitle = {arXiv preprint arXiv:2308.08155},
    shorttitle = {{AutoGen}},
    url = {https://www.semanticscholar.org/paper/AutoGen%3A-Enabling-Next-Gen-LLM-Applications-via-Wu-Bansal/9ea0757c750ab1222a7442d3485a74d1c526b04c},
    urldate = {2026-08-13},
    author = {Wu, Qingyun and Bansal, Gagan and Zhang, Jieyu and Wu, Yiran and Li, Beibin and Zhu, E. and Jiang, Li and Zhang, Xiaoyun and Zhang, Shaokun and Liu, Jiale and Awadallah, A. and White, Ryen W. and Burger, D. and Wang, Chi},
    month = aug,
    year = {2023},
}

@article{singhProgPromptGeneratingSituated2023,
    address = {London, United Kingdom},
    title = {{ProgPrompt}: {Generating} {Situated} {Robot} {Task} {Plans} using {Large} {Language} {Models}},
    copyright = {https://doi.org/10.15223/policy-029},
    shorttitle = {{ProgPrompt}},
    url = {https://ieeexplore.ieee.org/document/10161317/},
    doi = {10.1109/ICRA48891.2023.10161317},
    urldate = {2026-08-13},
    journal = {2023 IEEE International Conference on Robotics and Automation (ICRA)},
    publisher = {IEEE},
    author = {Singh, Ishika and Blukis, Valts and Mousavian, Arsalan and Goyal, Ankit and Xu, Danfei and Tremblay, Jonathan and Fox, Dieter and Thomason, Jesse and Garg, Animesh},
    month = may,
    year = {2023},
    pages = {11523--11530},
}

@article{wuChat2SVGVectorGraphics2025,
    address = {Nashville, TN, USA},
    title = {{Chat2SVG}: {Vector} {Graphics} {Generation} with {Large} {Language} {Models} and {Image} {Diffusion} {Models}},
    copyright = {https://doi.org/10.15223/policy-029},
    shorttitle = {{Chat2SVG}},
    url = {https://ieeexplore.ieee.org/document/11092564/},
    doi = {10.1109/CVPR52734.2025.02206},
    urldate = {2026-08-13},
    journal = {2025 IEEE/CVF Conference on Computer Vision and Pattern Recognition (CVPR)},
    publisher = {IEEE},
    author = {Wu, Ronghuan and Su, Wanchao and Liao, Jing},
    month = jun,
    year = {2025},
    pages = {23690--23700},
}

@article{thierryExtensionsWitnessMethod2011,
    series = {Solid and {Physical} {Modeling} 2010},
    title = {Extensions of the witness method to characterize under-, over- and well-constrained geometric constraint systems},
    volume = {43},
    issn = {0010-4485},
    url = {https://www.sciencedirect.com/science/article/pii/S0010448511001606},
    doi = {10.1016/j.cad.2011.06.018},
    number = {10},
    urldate = {2026-08-13},
    journal = {Computer-Aided Design},
    author = {Thierry, Simon E. B. and Schreck, Pascal and Michelucci, Dominique and Fünfzig, Christoph and Génevaux, Jean-David},
    month = oct,
    year = {2011},
    pages = {1234--1249},
}

@inproceedings{fanTraceCADTraceGuidedRepair2026,
    title = {{TraceCAD}: {Trace}-{Guided} {Repair} for {Agentic} {CAD} {Generation}},
    shorttitle = {{TraceCAD}},
    url = {https://www.semanticscholar.org/paper/TraceCAD%3A-Trace-Guided-Repair-for-Agentic-CAD-Fan-Ni/b06fff0aa9a211df31e448d4548c02d324c84d22},
    urldate = {2026-08-13},
    author = {Fan, Fengxiao and Ni, Jingzhe and Sang, Fan and Yin, Xiaolong and Liu, Yu and Tong, Ruofeng and Tang, Min and Du, Peng},
    month = aug,
    year = {2026},
}

@article{huIterCADIterativeMultimodal2026,
    title = {{IterCAD}: {An} {Iterative} {Multimodal} {Agent} for {Visually}-{Grounded} {CAD} {Generation} and {Editing}},
    copyright = {Creative Commons Attribution Non Commercial Share Alike 4.0 International},
    shorttitle = {{IterCAD}},
    url = {https://arxiv.org/abs/2606.13368},
    doi = {10.48550/ARXIV.2606.13368},
    urldate = {2026-08-13},
    publisher = {arXiv},
    author = {Hu, Tao and Ai, Jiaxin and Wen, Licheng and Li, Xueheng and Zou, Shu and Li, Siqi and Deng, Nianchen and Cai, Xinyu and Zhou, Hongbin and Cai, Pinlong and Fu, Daocheng and Yang, Yu and Zhang, Hairong and Shi, Botian and Yang, Xuemeng},
    month = jul,
    year = {2026},
}

@article{liuEmbodiedCADSolverGrounded2026,
    title = {Embodied {CAD}: {Solver}-{Grounded} {LLM} {Agents} for {Parametric} {B}-{Rep} {Assembly} {Modeling}},
    copyright = {arXiv.org perpetual, non-exclusive license},
    shorttitle = {Embodied {CAD}},
    url = {https://arxiv.org/abs/2606.31252},
    doi = {10.48550/ARXIV.2606.31252},
    urldate = {2026-08-13},
    publisher = {arXiv},
    author = {Liu, Fumin and Zhou, Haoyu and Hao, Fei and Yang, Lin},
    month = jul,
    year = {2026},
}

@misc{qwenQwen25TechnicalReport2025,
    title = {Qwen2.5 {Technical} {Report}},
    url = {http://arxiv.org/abs/2412.15115},
    doi = {10.48550/arXiv.2412.15115},
    urldate = {2026-08-13},
    publisher = {arXiv},
    author = {Qwen and Yang, An and Yang, Baosong and Zhang, Beichen and Hui, Binyuan and others},
    month = jan,
    year = {2025},
    note = {arXiv:2412.15115 [cs.CL]},
}

@misc{nandaEmergentLinearRepresentations2023,
    title = {Emergent {Linear} {Representations} in {World} {Models} of {Self}-{Supervised} {Sequence} {Models}},
    url = {http://arxiv.org/abs/2309.00941},
    doi = {10.48550/arXiv.2309.00941},
    urldate = {2026-08-13},
    publisher = {arXiv},
    author = {Nanda, Neel and Lee, Andrew and Wattenberg, Martin},
    month = sep,
    year = {2023},
    note = {arXiv:2309.00941 [cs.LG]},
}

@misc{gurneeLanguageModelsRepresent2024,
    title = {Language {Models} {Represent} {Space} and {Time}},
    url = {http://arxiv.org/abs/2310.02207},
    doi = {10.48550/arXiv.2310.02207},
    urldate = {2026-08-13},
    publisher = {arXiv},
    author = {Gurnee, Wes and Tegmark, Max},
    month = mar,
    year = {2024},
    note = {arXiv:2310.02207 [cs.LG]},
}

@misc{liEmergentWorldRepresentations2024,
    title = {Emergent {World} {Representations}: {Exploring} a {Sequence} {Model} {Trained} on a {Synthetic} {Task}},
    shorttitle = {Emergent {World} {Representations}},
    url = {http://arxiv.org/abs/2210.13382},
    doi = {10.48550/arXiv.2210.13382},
    urldate = {2026-08-13},
    publisher = {arXiv},
    author = {Li, Kenneth and Hopkins, Aspen K. and Bau, David and Viégas, Fernanda and Pfister, Hanspeter and Wattenberg, Martin},
    month = jun,
    year = {2024},
    note = {arXiv:2210.13382 [cs.LG]},
}

@misc{parkLinearRepresentationHypothesis2024,
    title = {The {Linear} {Representation} {Hypothesis} and the {Geometry} of {Large} {Language} {Models}},
    url = {http://arxiv.org/abs/2311.03658},
    doi = {10.48550/arXiv.2311.03658},
    urldate = {2026-08-13},
    publisher = {arXiv},
    author = {Park, Kiho and Choe, Yo Joong and Veitch, Victor},
    month = jul,
    year = {2024},
    note = {arXiv:2311.03658 [cs.CL]},
}

@misc{wietingNoTrainingRequired2019,
    title = {No {Training} {Required}: {Exploring} {Random} {Encoders} for {Sentence} {Classification}},
    shorttitle = {No {Training} {Required}},
    url = {http://arxiv.org/abs/1901.10444},
    doi = {10.48550/arXiv.1901.10444},
    urldate = {2026-08-13},
    publisher = {arXiv},
    author = {Wieting, John and Kiela, Douwe},
    month = jan,
    year = {2019},
    note = {arXiv:1901.10444 [cs.CL]},
}

@inproceedings{hewittConditionalProbingMeasuring2021,
    address = {Online and Punta Cana, Dominican Republic},
    title = {Conditional probing: measuring usable information beyond a baseline},
    shorttitle = {Conditional probing},
    url = {https://aclanthology.org/2021.emnlp-main.122/},
    doi = {10.18653/v1/2021.emnlp-main.122},
    urldate = {2026-08-13},
    booktitle = {Proceedings of the 2021 {Conference} on {Empirical} {Methods} in {Natural} {Language} {Processing}},
    publisher = {Association for Computational Linguistics},
    author = {Hewitt, John and Ethayarajh, Kawin and Liang, Percy and Manning, Christopher},
    editor = {Moens, Marie-Francine and Huang, Xuanjing and Specia, Lucia and Yih, Scott Wen-tau},
    month = nov,
    year = {2021},
    pages = {1626--1639},
}

@inproceedings{ravichanderProbingProbingParadigm2021,
    address = {Online},
    title = {Probing the {Probing} {Paradigm}: {Does} {Probing} {Accuracy} {Entail} {Task} {Relevance}?},

    booktitle = {Proceedings of the 16th {Conference} of the {European} {Chapter} of the {Association} for {Computational} {Linguistics}: {Main} {Volume}},
    publisher = {Association for Computational Linguistics},
    author = {Ravichander, Abhilasha and Belinkov, Yonatan and Hovy, Eduard},
    editor = {Merlo, Paola and Tiedemann, Jorg and Tsarfaty, Reut},
    month = apr,
    year = {2021},
    pages = {3363--3377},
}

@misc{turnerSteeringLanguageModels2024,
    title = {Steering {Language} {Models} {With} {Activation} {Engineering}},
    url = {http://arxiv.org/abs/2308.10248},
    doi = {10.48550/arXiv.2308.10248},
    urldate = {2026-08-13},
    publisher = {arXiv},
    author = {Turner, Alexander Matt and Thiergart, Lisa and Leech, Gavin and Udell, David and Vazquez, Juan J. and Mini, Ulisse and MacDiarmid, Monte},
    month = oct,
    year = {2024},
    note = {arXiv:2308.10248 [cs.CL]},
}

@misc{zouRepresentationEngineeringTopDown2025,
    title = {Representation {Engineering}: {A} {Top}-{Down} {Approach} to {AI} {Transparency}},
    shorttitle = {Representation {Engineering}},
    url = {http://arxiv.org/abs/2310.01405},
    doi = {10.48550/arXiv.2310.01405},
    urldate = {2026-08-13},
    publisher = {arXiv},
    author = {Zou, Andy and Phan, Long and Chen, Sarah and Campbell, James and Guo, Phillip and Ren, Richard and Pan, Alexander and Yin, Xuwang and Mazeika, Mantas and Dombrowski, Ann-Kathrin and Goel, Shashwat and Li, Nathaniel and Byun, Michael J. and Wang, Zifan and Mallen, Alex and Basart, Steven and Koyejo, Sanmi and Song, Dawn and Fredrikson, Matt and Kolter, J. Zico and Hendrycks, Dan},
    month = mar,
    year = {2025},
    note = {arXiv:2310.01405 [cs.LG]},
}

@misc{willisFusion360Gallery2021,
    title = {Fusion 360 {Gallery}: {A} {Dataset} and {Environment} for {Programmatic} {CAD} {Construction} from {Human} {Design} {Sequences}},
    shorttitle = {Fusion 360 {Gallery}},
    url = {http://arxiv.org/abs/2010.02392},
    doi = {10.48550/arXiv.2010.02392},
    language = {en},
    urldate = {2026-08-17},
    publisher = {arXiv},
    author = {Willis, Karl D. D. and Pu, Yewen and Luo, Jieliang and Chu, Hang and Du, Tao and Lambourne, Joseph G. and Solar-Lezama, Armando and Matusik, Wojciech},
    month = may,
    year = {2021},
    note = {arXiv:2010.02392 [cs.LG]},
}

@inproceedings{zhangLanguageModelingTeaches2018,
  address = {Brussels, Belgium},
  title = {Language {Modeling} {Teaches} {You} {More} than {Translation} {Does}: {Lessons} {Learned} {Through} {Auxiliary} {Syntactic} {Task} {Analysis}},
  shorttitle = {Language {Modeling} {Teaches} {You} {More} than {Translation} {Does}},
  url = {https://aclanthology.org/W18-5448/},
  doi = {10.18653/v1/W18-5448},
  urldate = {2026-10-27},
  booktitle = {Proceedings of the 2018 {EMNLP} {Workshop} {BlackboxNLP}: {Analyzing} and {Interpreting} {Neural} {Networks} for {NLP}},
  publisher = {Association for Computational Linguistics},
  author = {Zhang, Kelly and Bowman, Samuel R.},
  editor = {Linzen, Tal and Chrupała, Grzegorz and Alishahi, Afra},
  month = nov,
  year = {2018},
  pages = {359--361},
}

@inproceedings{zhaoCalibrateUseImproving2021,
  title = {Calibrate {Before} {Use}: {Improving} {Few}-shot {Performance} of {Language} {Models}},
  issn = {2640-3498},
  shorttitle = {Calibrate {Before} {Use}},
  url = {https://proceedings.mlr.press/v139/zhao21c.html},
  language = {en},
  urldate = {2026-10-27},
  booktitle = {Proceedings of the 38th {International} {Conference} on {Machine} {Learning}},
  publisher = {PMLR},
  author = {Zhao, Zihao and Wallace, Eric and Feng, Shi and Klein, Dan and Singh, Sameer},
  month = jul,
  year = {2021},
  pages = {12697--12706},
}

@misc{qwen3.5,
    title  = {{Qwen3.5}: Towards Native Multimodal Agents},
    author = {{Qwen Team}},
    month  = {February},
    year   = {2026},
    url    = {https://qwen.ai/blog?id=qwen3.5}
}

@misc{olmoOlmo32026,
  title = {Olmo 3},
  url = {http://arxiv.org/abs/2512.13961},
  doi = {10.48550/arXiv.2512.13961},
  urldate = {2026-10-27},
  publisher = {arXiv},
  author = {Olmo, Team and Ettinger, Allyson and Bertsch, Amanda and Kuehl, Bailey and Graham, David and Heineman, David and Groeneveld, Dirk and Brahman, Faeze and Timbers, Finbarr and Ivison, Hamish and Morrison, Jacob and Poznanski, Jake and Lo, Kyle and Soldaini, Luca and Jordan, Matt and Chen, Mayee and Noukhovitch, Michael and Lambert, Nathan and Walsh, Pete and Dasigi, Pradeep and Berry, Robert and Malik, Saumya and Shah, Saurabh and Geng, Scott and Arora, Shane and Gupta, Shashank and Anderson, Taira and Xiao, Teng and Murray, Tyler and Romero, Tyler and Graf, Victoria and Asai, Akari and Bhagia, Akshita and Wettig, Alexander and Liu, Alisa and Rangapur, Aman and Anastasiades, Chloe and Huang, Costa and Schwenk, Dustin and Trivedi, Harsh and Magnusson, Ian and Lochner, Jaron and Liu, Jiacheng and Miranda, Lester James V. and Sap, Maarten and Morgan, Malia and Schmitz, Michael and Guerquin, Michal and Wilson, Michael and Huff, Regan and Bras, Ronan Le and Xin, Rui and Shao, Rulin and Skjonsberg, Sam and Shen, Shannon Zejiang and Li, Shuyue Stella and Wilde, Tucker and Pyatkin, Valentina and Merrill, Will and Chang, Yapei and Gu, Yuling and Zeng, Zhiyuan and Sabharwal, Ashish and Zettlemoyer, Luke and Koh, Pang Wei and Farhadi, Ali and Smith, Noah A. and Hajishirzi, Hannaneh},
  month = apr,
  year = {2026},
  note = {arXiv:2512.13961 [cs.CL]},
}
\bibliographystyle{tmlr}

\appendix
% \documentclass[letterpaper]{article}
% \usepackage[submission]{aaai2026}
% \usepackage{times}
% \usepackage{helvet}
% \usepackage{courier}
% \usepackage[hyphens]{url}
% \usepackage{graphicx}
% \urlstyle{rm}
% \def\UrlFont{\rm}
% \usepackage{natbib}
% \usepackage{caption}
% \frenchspacing
% \setlength{\pdfpagewidth}{8.5in}
% \setlength{\pdfpageheight}{11in}
% \usepackage{amsmath,amssymb,booktabs}  % 按你正文实际用到的包补全
% % \usepackage{xr-hyper}
% % \externaldocument{paper}

% \title{Encoded but Not Actionable: Technical Appendix}
% \begin{document}
% \maketitle
\clearpage
\appendix

\section{Task and Label Details}
\label{app:task_details}

\subsection{P1 and P2 Label Sets}
\label{app:constraint_gallery}

Figure~\ref{fig:constraint_gallery} illustrates the eight P1 classes:
seven pairwise \texttt{EdgeOp} relations and a sampled
\textsc{NoConstraint} class. It also shows the three P2 labels:
under-, well-, and over-constrained.

\begin{figure}[h]
\centering
\includegraphics[width=\columnwidth]{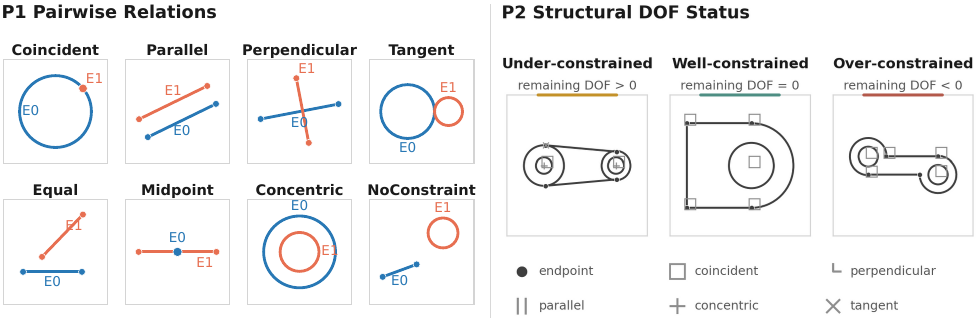}
\caption{P1 pairwise constraint classes and P2 structural DOF labels in
SketchGraphs. Highlighted entities indicate the pair probed in P1;
\texttt{EdgeOp} tokens are excluded from the LLM input.}
\label{fig:constraint_gallery}
\end{figure}
%% NOTE: figure above still shows Distance/Angle/Symmetric -- must be regenerated to match the seven classes listed above.

\subsection{P2 Label Quality Checks}
\label{app:p2quality}
\paragraph{Constraint feasibility.}
We sample 2,000 sketches from each P2 class and test whether their
constraints can be satisfied simultaneously. The solver returns a
checkable result for 1,932 under-constrained, 1,992 well-constrained,
and 1,989 over-constrained sketches. Among these, the satisfiable proportions are
99.84\% (95\% CI:
99.55--99.97), 99.95\% (99.72--100.00), and 99.70\%
(99.34--99.89), respectively. These results
suggest that contradictory constraints are rare, but satisfiability
alone does not validate the DOF-status labels: redundant constraints
may be jointly satisfiable while removing fewer
independent degrees of freedom than assumed by the heuristic.

\section{Probe Robustness Checks}
\label{app:probe_}
\subsection{Cross-Dataset P1 Check}
\label{app:fusion360}

We repeat the P1 probing analysis on Fusion~360 Gallery
reconstruction data (r1.0.1) using three representative
architectures. Across models, pretrained P1 macro-F1 peaks between
0.640 and 0.655. At the same pretrained-selected layers, the
corresponding random-init scores range from 0.486 to 0.523, yielding
pretraining gains of 0.118--0.160 (Table~\ref{tab:fusion360}).
Selectivity remains high at 0.510--0.528. These results show that P1
decodability and its improvement from pretraining extend beyond
SketchGraphs. Because our current Fusion~360 pipeline does not construct P2 labels
with the \texttt{get\_sequence\_dof} heuristic used for SketchGraphs,
this analysis tests only the cross-dataset robustness of P1.

\begin{table}[H]
\caption{Cross-dataset P1 results on Fusion~360 Gallery. Random-init
scores are measured at the pretrained model's selected layer
$\ell^*$. Gain is pretrained minus random-init macro-F1; selectivity
is pretrained minus shuffled-label performance.}
\label{tab:fusion360}

\begin{center}
\small
\setlength{\tabcolsep}{5pt}
\begin{tabular}{@{}lccccc@{}}
\toprule
Model
& $\ell^*/L$
& Pretrained P1 [95\% CI]
& Random-init P1
& Gain
& Selectivity \\
\midrule
Qwen2.5-3B
& 26/36 & .646 [.625, .666] & .486 & .160 & .510 \\
Llama-3.1-8B
& 14/32 & .655 [.636, .672] & .523 & .132 & .528 \\
OLMo-3-7B
& 18/32 & .640 [.621, .657] & .523 & .118 & .513 \\
\bottomrule
\end{tabular}
\end{center}
\end{table}
\subsection{Chance-Normalized Dissociation Index}
\label{app:di_norm}

Because P1 and P2 have different uniform-class reference levels
(\(1/8\) and \(1/3\), respectively), we normalize each task's
pretrained-over-random gain by its available headroom:
\begin{equation}
\label{eq:di_norm}
\begin{aligned}
\mathrm{DI}_{\mathrm{norm}}
={}&
\frac{
F_{1,\mathrm{pre}}^{P1}(\ell^*)
-
F_{1,\mathrm{rand}}^{P1}(\ell^*)
}{
1-\frac{1}{8}
}
&-
\frac{
F_{1,\mathrm{pre}}^{P2}(\ell^*)
-
F_{1,\mathrm{rand}}^{P2}(\ell^*)
}{
1-\frac{1}{3}
}.
\end{aligned}
\end{equation}

\begin{table}[H]
\caption{Chance-normalized DI at each model's pretrained
P1-selected layer $\ell^*$.}
\label{tab:normalized-di}

\centering
\setlength{\tabcolsep}{8pt}
\begin{tabular}{@{}lcc@{}}
\toprule
Model & $\ell^*$ & $\mathrm{DI}_{\mathrm{norm}}$ \\
\midrule
Qwen2.5-0.5B & 14 & .151 \\
Qwen2.5-1.5B & 16 & .171 \\
Qwen2.5-3B   & 26 & .174 \\
Qwen2.5-7B   & 20 & .159 \\
Qwen3.5-9B   & 14 & .146 \\
Llama-3.1-8B & 14 & .143 \\
Mistral-7B   & 14 & .130 \\
OLMo-3-7B    & 14 & .178 \\
\bottomrule
\end{tabular}
\end{table}

Chance-normalized DI remains positive for all eight models
(0.130--0.178; Table~\ref{tab:normalized-di}), showing that the
dissociation is not explained by the tasks' different chance levels.
\section{P3 Generation Diagnostics}
\label{app:p3_diagnostics}

\begin{table}[H]
\caption{P3 first-token forced-choice accuracy conditional on the
true constraint class for eight models. Each balanced test set
contains 2,000 examples, with 250 per class. Bold marks the highest
class accuracy for each model; dashes denote exact zeros.}
\label{tab:p3-per-class}

\centering
\setlength{\tabcolsep}{5pt}
\begin{tabular}{@{}lcccccccc@{}}
\toprule
Model
& Coin & Par & Perp & Tan & Eq & Mid & Conc & NoC \\
\midrule
Qwen2.5-0.5B
& .260 & .136 & .076 & \textbf{.932}
& -- & -- & -- & -- \\

Qwen2.5-1.5B
& .012 & -- & .364 & \textbf{.948}
& -- & -- & -- & .004 \\

Qwen2.5-3B
& .412 & -- & .092 & \textbf{.956}
& -- & -- & .008 & -- \\

Qwen2.5-7B
& .168 & .804 & .312 & \textbf{.956}
& -- & -- & .516 & .008 \\

Qwen3.5-9B
& .200 & .464 & .788 & \textbf{.916}
& .112 & .052 & .628 & .068 \\

Mistral-7B
& \textbf{.996} & -- & -- & --
& -- & -- & -- & -- \\

Llama-3.1-8B
& .240 & .072 & .004 & .312
& .008 & -- & .112 & \textbf{.804} \\

OLMo-3-7B
& -- & .068 & .680 & .120
& -- & -- & -- & \textbf{.928} \\
\bottomrule
\end{tabular}
\end{table}

To measure output-label preferences without geometric content, we
evaluate ten content-free prompt variants. For each model, we average
the first-token probabilities assigned to each label across the ten
prompts and normalize the resulting probabilities over the eight
candidate labels. The resulting priors are strongly
architecture-specific: the Qwen2.5 models favor
\textsc{Tangent}, Mistral and Llama favor \textsc{Coincident}, and
Qwen3.5 and OLMo favor \textsc{NoConstraint}
(Table~\ref{tab:p3-calibration}).

\begin{table}[H]
\caption{Label-normalized mean first-token probabilities across ten
content-free prompt variants, followed by raw and content-free
calibrated P3 macro-F1. Bold marks the largest content-free prior for
each model.}
\label{tab:p3-calibration}

\centering
\setlength{\tabcolsep}{4pt}
\begin{tabular}{@{}lcccccccccc@{}}
\toprule
Model
& Coin & Par & Perp & Tan & Eq & Mid & Conc & NoC
& Raw P3 & Cal.\ P3 \\
\midrule
Qwen2.5-0.5B
& .191 & .159 & .173 & \textbf{.255}
& .093 & .011 & .005 & .114
& .095 & .144 \\

Qwen2.5-1.5B
& .108 & .066 & .243 & \textbf{.267}
& .081 & .023 & .069 & .143
& .072 & .142 \\

Qwen2.5-3B
& .239 & .047 & .083 & \textbf{.367}
& .121 & .035 & .005 & .104
& .089 & .259 \\

Qwen2.5-7B
& .115 & .232 & .138 & \textbf{.242}
& .031 & .036 & .062 & .144
& .257 & .298 \\

Qwen3.5-9B
& .183 & .104 & .195 & .120
& .048 & .047 & .088 & \textbf{.215}
& .344 & .380 \\

Mistral-7B
& \textbf{.329} & .149 & .097 & .053
& .075 & .031 & .052 & .215
& .028 & .152 \\

Llama-3.1-8B
& \textbf{.254} & .104 & .110 & .114
& .092 & .057 & .043 & .227
& .141 & .265 \\

OLMo-3-7B
& .055 & .055 & .116 & .195
& .045 & .053 & .004 & \textbf{.477}
& .129 & .239 \\
\bottomrule
\end{tabular}
\end{table}

\FloatBarrier
\section{Intervention Details}
\label{app:steering}

We evaluate steering at both a common early layer, \(L=4\), and each
model's P1 probe readout layer \(\ell^*\). For P1, the intervention
adds a class mean-difference direction to both queried entity spans;
for P2, it is applied to all entity spans. We test
\(\alpha\in\{0.5,1,2,4,8\}\) on 200 examples per task and compare each
mean-difference direction with 10 matched random directions.

Under first-token scoring, P2 produces no target-class flips in any
tested condition. For P1, target-class flip rates at \(\alpha=8\)
range from 0 to 0.070 and do not consistently exceed matched
random-direction controls (0.005--0.082). Table~\ref{tab:steering_full}
reports unrestricted label-change rates at \(\alpha=8\). Larger
interventions can change many predictions, but these changes are not
reliably directed toward the intended class. Thus, restoring an
example-specific clean activation can repair a probe prediction,
whereas adding a single global class direction does not provide
reliable targeted control.

\begin{table}[H]
\caption{Fraction of predictions changed to any label at
$\alpha=8$ under first-token scoring. Random-direction values are
averaged over ten matched directions.}
\label{tab:steering_full}

\centering
\setlength{\tabcolsep}{8pt}
\begin{tabular}{@{}llcccc@{}}
\toprule
& & \multicolumn{2}{c}{P1} & \multicolumn{2}{c}{P2} \\
\cmidrule(lr){3-4}
\cmidrule(lr){5-6}
Model & Layer
& Mean diff. & Random
& Mean diff. & Random \\
\midrule
Qwen2.5-3B
& L4       & .050 & .023 & .000 & .001 \\
& $\ell^*$ & .010 & .009 & .000 & .000 \\
\addlinespace

Llama-3.1-8B
& L4       & .875 & .321 & .335 & .370 \\
& $\ell^*$ & .145 & .065 & .090 & .100 \\
\addlinespace

OLMo-3-7B
& L4       & .285 & .193 & .995 & .507 \\
& $\ell^*$ & .080 & .048 & .955 & .219 \\
\bottomrule
\end{tabular}
\end{table}

% \end{document}
\end{document}